\documentclass[11pt]{article}
\usepackage{booktabs}
\usepackage{longtable}
\usepackage{amsmath}
\usepackage{booktabs}
\usepackage{pifont} 

\usepackage[most]{tcolorbox}
\usepackage{amsmath}

\definecolor{boxblue}{RGB}{41, 128, 185}
\definecolor{textgray}{RGB}{44, 62, 80}
\definecolor{successgreen}{RGB}{39, 174, 96}

\newtcolorbox{TranslationBox}[1]{%
    colback=white,
    colframe=boxblue,
    arc=3pt,
    outer arc=3pt,
    boxrule=0.5pt,
    leftrule=4pt, 
    title={\textbf{#1}},
    coltitle=white,
    attach boxed title to top left={yshift=-2mm, xshift=2mm},
    boxed title style={colback=boxblue, sharp corners=south},
    fonttitle=\small\sffamily,
    fontupper=\small\sffamily\color{textgray},
    boxsep=2pt,
    top=6pt,
    bottom=6pt,
    left=6pt,
    right=6pt
}

\usepackage[final]{acl}
\usepackage[capitalize,noabbrev]{cleveref}
\crefname{section}{Sec.}{Secs.}
\Crefname{section}{Sec.}{Secs.}
\crefname{figure}{Fig.}{Figs.}
\Crefname{figure}{Fig.}{Figs.}
\crefname{table}{Tab.}{Tabs.}
\Crefname{table}{Tab.}{Tabs.}
\crefname{appendix}{App.}{Apps.}
\Crefname{appendix}{App.}{Apps.}
\usepackage{float}
\usepackage{times}
\usepackage{latexsym}
\usepackage[utf8]{inputenc}
\usepackage[T2A]{fontenc}
\usepackage[russian,english]{babel}
\usepackage{multirow}
\usepackage[most]{tcolorbox}
\usepackage{xcolor} 
\usepackage{amssymb}
\usepackage[most]{tcolorbox} 
\tcbuselibrary{skins, breakable}

\usepackage[T1]{fontenc}
\usepackage{fancyvrb}

\usepackage[utf8]{inputenc}

\usepackage{microtype}

\usepackage{inconsolata}

\usepackage{graphicx}
\usepackage{url}
\usepackage{multirow}

\usepackage{fontspec}
\newfontfamily\kalpurush{kalpurush.ttf}

\newcommand{\bn}[1]{{\kalpurush #1}}
\newcommand{\DataReleaseURL}[0]{\url{https://github.com/SharifMAbdullah/breaking-the-silence}}
\usepackage{inconsolata}
\usepackage{import}
\usepackage{makecell}

\usepackage[scientific-notation=true]{siunitx}
\usepackage{tabularx}
\usepackage[linesnumbered,ruled,vlined]{algorithm2e}
\usepackage{amsmath} 
\usepackage{algpseudocode}
\usepackage{booktabs}   
\usepackage{graphicx}   
\usepackage{enumitem}
\usepackage{tablefootnote}

\title{Breaking the Silence: A Dataset and Benchmark for Bangla Text-to-Gloss Translation}

\author{
  \textbf{Sharif Mohammad Abdullah}\textsuperscript{1},
  \textbf{Abhijit Paul}\textsuperscript{1},
  \textbf{Shubhashis Roy Dipta}\textsuperscript{2},\\[0.3em]
  \textbf{Zarif Masud}\textsuperscript{3},
  \textbf{Shebuti Rayana}\textsuperscript{4},
  \textbf{Ahmedul Kabir}\textsuperscript{1}
  \\[0.8em]
  \textsuperscript{1}IIT, University of Dhaka, Bangladesh \quad
  \textsuperscript{2}University of Maryland, Baltimore County, USA \\[0.3em]
  \textsuperscript{3}Toronto Metropolitan University, Canada \quad \textsuperscript{4}SUNY, Old Westbury, USA
  \\[0.4em]
  \small{\textbf{Correspondence:} \href{mailto:bsse1211@iit.du.ac.bd}{bsse1211@iit.du.ac.bd}}}

\begin{document}
\maketitle
\begin{abstract} 
Gloss is a written approximation that bridges Sign Language (SL) and its corresponding spoken language. Despite a deaf and hard-of-hearing population of at least 3 million in Bangladesh, Bangla Sign Language (BdSL) remains largely understudied, with very limited prior work on Bangla text-to-gloss translation and no publicly available parallel dataset at the time of this work. To address this gap, we construct the first publicly available Bangla text-to-gloss dataset, consisting of 1,000 manually annotated and 4,000 synthetically generated Bangla sentence–gloss pairs, along with 159 expert human-annotated pairs used as a test set.
Our experimental framework performs a comparative analysis between several fine-tuned open-source models and a leading closed-source LLM to evaluate their performance in low-resource BdSL translation. GPT-5.4 achieves the best automatic metric scores, while open source models perform competitively despite being much smaller.
This work introduces the first publicly available expert-annotated Bangla text-to-gloss dataset, benchmark and trained model for Bangla text-to-gloss translation. It also demonstrates the effectiveness of systematically generated synthetic data for addressing challenges in low-resource sign language translation.\footnote{\DataReleaseURL}

\end{abstract}

\section{Introduction}
Sign language has been the primary method of communication for deaf people for centuries \cite{cantin2022perspectives}. But deaf people often use lip reading, text-based communication \cite{barnett2002communication} to interact with others. Sign Language Generation (SLG), the process of converting text into sign, can improve the quality of life of deaf people by helping them receive information in their preferred language. In addition, the deaf community has expressed their preference in using sign over other media \cite{middleton2010preferences}.

In the context of Sign Language (SL), gloss is considered as the written approximation of SL which uses words as ``labels'' for each sign along with various grammatical notes. A sign can have multiple meanings depending on the context of the sentence. For example, the sign RIGHT can mean a direction, a correct answer, or a legal/moral entitlement. "Your answer is correct." would be translated as "YOUR ANSWER RIGHT" and "Turn right at the light." would be translated to "TRAFFIC-LIGHT FORWARD TURN-RIGHT", even though the sign is same.

 \begin{figure}[!t]
  \includegraphics[width=\linewidth]{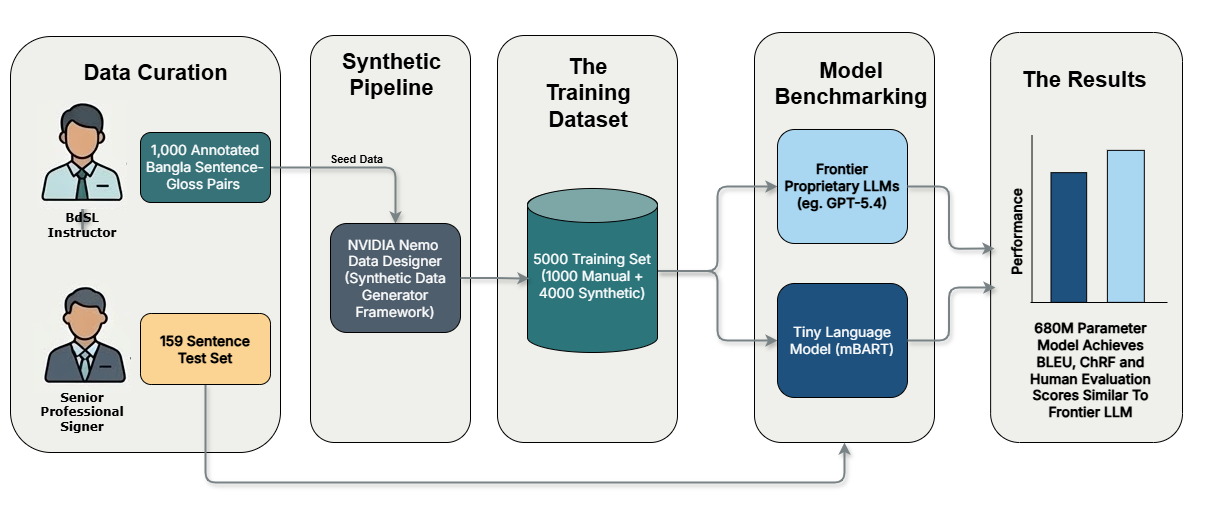}
  \caption{Our T2G pipeline: 1,000 expert-annotated pairs seed an LLM-based data generator to produce 4,000 synthetic sentences, for a total of 5,000 sentences. Fine-tuned models are evaluated against a proprietary LLM baseline using both automatic metrics and human judgment.}
  \label{fig:stltask}
\end{figure}

Bangla, a language spoken by 272.7 million individuals \cite{SULTANA2025111276}, also has a substantial deaf community that uses the Bangla Sign Language daily. According to the government defined disability categories, among the people of Bangladesh, hearing disability is at 0.29\%~\cite{statbbs}, although \citet{alauddin2004deafness} suggest figures as high as 3 million.
However, BdSL is a low-resource language. In our work, we introduce a novel dataset representing Bangla sentences and their gloss form. \citet{gomez2021syntax} shows that the use of glosses as intermediaries improves the translation process. 
To address this gap, we present following contributions, 
\begin{itemize}[left=1pt]
    \item We curated a dataset of 1,000 hand-annotated Bangla sentences and their corresponding glosses, annotated by a certified BdSL instructor. To ensure a rigorous evaluation, a separate test set of 159 sentences was independently developed by a senior professional BdSL signer.
    \item We generate 4000 synthetic gloss with \citet{nemo-data-designer} framework using the manually labeled data as the seed for the generation process
    \item We benchmark a diverse set of proprietary and open-source models to evaluate their performance on our dataset, and show that small models finetuned on our dataset can achieve performance comparable to proprietary LLM
\end{itemize}
 
Our dataset can serve as the foundation for the text-to-gloss-to-pose-to-video pipeline \cite{moryossef2023open}. Our results suggest that LLM-based synthetic data generation, seeded with a small expert-annotated dataset, can serve as a cost-effective complement to manual annotation for low-resource sign language tasks.
\section{Literature Review}

\paragraph{Text-to-Gloss.}
Text-to-gloss is a necessary task for sign language generation (SLG). Additionally, for tasks like text-to-sign, text-to-gloss translation is a necessary step. \citet{moryossef2023open} demonstrates that T2G is a crucial part in the text-to-gloss-to-pose-to-video pipeline. Despite such a necessity, text-to-gloss remains a low-resource domain.
\citet{gomez2021syntax} propose a syntax-aware Transformer for English text-to-gloss translation, showing that incorporating syntactic information improves gloss generation quality. On the data side, \citet{moryossef2021data} address the scarcity of parallel gloss corpora by proposing rule-based heuristics that generate pseudo-parallel gloss-text pairs from monolingual spoken language text, demonstrating that pre-training on such synthetic data improves gloss translation by up to 3.14 BLEU. Our work builds on this line of research by applying LLM-based synthetic generation to BdSL, a language pair not previously studied.

\paragraph{Bangla Sign Language.}
\citet{moryossef2021slp} describes 20 distinct tasks for sign language translation. Among them, BdSL mainly focuses on the video (sign) to text task \cite{zeeon2024btvsl, 10306914, 9456176, alam2021two, 7760123, 9074370}. Recent works include BdSLW60 prepared by \citet{rubaiyeat2025bdslw60}. This is an unconstrained word-level video dataset of 60 signs across over 9,000 trials. Building upon this work, \citet{rubaiyeat2025bdslw401} scaled word recognition with BdSLW401 containing over 102,000 video samples. Another work by \citet{rubaiyeat2025bangla} pioneered IsharaKhobor, a sentence translation dataset harvested from TV news.
Few works have focused on the text-to-sign and text-to-pose tasks \cite{shahriar2017communication, sarkar2009translator, 7760123}. \citet{islam2025isharakotha} introduced IsharaKotha, the first-ever HamNoSys-based BdSL corpus of 3,823 words. However, Bangla lacks extensive research in gloss annotation, though recent work by \citet{saha2026introducing} introduced Bangla-SGP, a parallel corpus of sentence-gloss pairs built with rule-based pipeline. In fact, there is a general scarcity of text-to-gloss datasets across languages~\citep{moryossef2021data}; we provide a comparison of existing corpora in \cref{app:sl_corpora}.

\paragraph{LLM for synthetic data generation.}
LLM-generated synthetic data has been shown to improve model performance when combined with human-labeled data, particularly under budget constraints~\cite{matta2024investigating, zhezherau2024hybrid}. However, smaller models tend to exhibit biases towards their own generated data, while larger models show more reliability~\cite{maheshwari2024efficacy}.

\section{Methodology}
\subsection{Field Study}
\label{sec:field_study}

We started by interviewing BdSL teachers and professional signers. Key findings: (i) there is no single standard for BdSL -- it varies between schools and individuals; (ii) BdSL follows rules of thumb rather than a strict grammar, such as using infinitive verb forms. Despite this, governmental and academic institutions have published standardization guides. We curated part of our test set from these resources and the rest with a professional sign language interpreter, who re-verified every sentence. A separate annotator prepared the training data to reflect natural variation. 
We chose two annotators for the training and the test sets to prevent overfitting and to generalize the model. If the same person wrote both the training and test data, a model could achieve a good score by copying that person's style. But this would not have meant the model learned BdSL. 
Annotator details are in \cref{appendix:annotator}.

\subsection{Constructing Bangla T2G dataset}
We construct a Bangla text-to-gloss (T2G) dataset which, to the best of our knowledge, is the first for Bangla language. We consider two data sources for Bangla T2G - human annotated data and synthetic data generated using LLM.

\subsubsection*{Curating hand annotated data}
All contributors provided informed consent. We provided the training annotator with 1,000 Bangla sentences covering diverse aspects of everyday life and asked them to translate each into its gloss form. This manually created data served as ground truth and as seed for synthetic data generation.

Our final dataset comprises 1,000 human-annotated and 4,000 synthetic training pairs, with a separate test set of 159 human-annotated sentences (\cref{app:dataset_table}).\label{table:dataset-combination}

\subsubsection*{Synthetic data generation using LLM}
LLM has been widely used for generating synthetic data in low-resource domain and has shown sufficient reliability \cite{zehady2026banglallama, alonso2024human}. To find whether LLM can be used reliably to generate BdSL gloss, we take a sample of 100 sentences from the manually annotated dataset to generate gloss form for 100 Bangla sentences with \citet{nemo-data-designer} framework and \texttt{GPT-4o} as the underlying LLM. In our prompt (provided in \cref{appendix:synth_prompt}), we mention the rules of thumb of BdSL, instruct the LLM to follow the patterns seen in the seed dataset and apply it all together to generate a new sentence-gloss pair for a new topic. We then evaluate these 100 text-to-gloss translations with our human annotator who created the training dataset who marked \textbf{75\%} of sentences as correct. After gaining satisfactory remarks on this pilot dataset from our annotator, we generate gloss representations for an additional 4000 Bangla sentences using the same framework. The pilot sentences are not included in this dataset.

We combine the above two data sources to construct our final dataset (full details in \cref{table:dataset-combination}), creating \textit{bangla-gloss}.

\begin{table*}[t]
\centering
\small
\begin{tabular}{@{}llcccc r@{}}
\toprule
\textbf{Model} & \textbf{Type} & \textbf{Parameters} & \textbf{BLEU} & \textbf{chrF} & \textbf{Human Eval. (n=100)} & \textbf{95\% CI} \\
\midrule
GPT-5.4     & Causal LM     & --          & \textbf{39.26} & \textbf{73.75} & \underline{68}\% & [58.3, 76.3] \\
Qwen-3       & Causal LM     & 8B        & 30.82 & 70.16 & \textbf{72}\%  & \textbf{[62.5, 79.9]} \\
Phi-4  \textsuperscript{$\dagger$}       & Causal LM     & 4B          & 2.26  & 26.95 & 0\% & [0.0, 3.7] \\
mBART-large & Seq2Seq       & 0.68B       & \underline{34.81} & \underline{70.81} & 65\% & [55.3, 73.6] \\
MarianMT\textsuperscript{$\dagger$}    & Seq2Seq       & 0.14B       & 7.11  & 44.60 & 2\% &[0.5, 7.0] \\
\bottomrule
\end{tabular}

\vspace{2pt}
{\small $\dagger$~Frequently produces misspelled tokens and incoherent gloss sequences, resulting in near-zero human acceptability.}

\caption{Automatic (BLEU-4, chrF) scores on the \textit{bangla-gloss} test set (159 sentences) and human evaluation scores (100 sentences). Proprietary models are prompted few-shot; open-source models are fine-tuned on our training data. Human evaluation reflects acceptability as judged by a certified BdSL signer (\cref{appendix:annotator}).}
\label{table:result}
\end{table*}


\subsection{Closed-source Baselines}
\texttt{GPT-5.4} represents the state-of-the-art in generative LLM. It was incorporated into this study to serve as a state-of-the-art baseline for few-shot translation capabilities.

\subsection{Trained Models for T2G Translation}
We fine-tune four open-source models on our proposed training set: (i)~\textbf{MarianMT}~\cite{tiedemann2004opus}, a dedicated NMT encoder-decoder pre-trained on OPUS; (ii)~\textbf{mBART-50}~\cite{liu2020multilingual}, a multilingual seq2seq model whose bidirectional encoder and autoregressive decoder are well-suited for low-resource translation; (iii)~\textbf{Qwen-3}~\cite{yang2025qwen3}, selected for its strong multilingual performance; and (iv)~\textbf{Phi-4} \cite{abdin2024phi}, included to assess whether a compact general-purpose LM can compete with dedicated translation models. Additional implementation details are provided in \cref{app:training}.

\subsection{Evaluation Scheme}
Following \citet{moryossef2021data}, we use three metrics - BLEU-4~\cite{papineni2002bleu}, chrF~\cite{popovic2015chrf} and Human Evaluation to evaluate the text-to-gloss translation models. These metrics provide insight into the surface-level similarity of generated outputs with reference glosses, with higher-order BLEU scores emphasizing longer, contiguous n-gram matches. We do not use METEOR as their authors state that Bangla is not a supported language\footnote{\url{https://www.cs.cmu.edu/~alavie/METEOR/}}. Although COMET supports Bangla, we do not use it because it hasn't seen BdSL in its training data, making it unsuitable for BdSL evaluation\footnote{\href{https://github.com/Unbabel/COMET?tab=readme-ov-file\#languages-covered}{\texttt{github.com/Unbabel/COMET}}}. Full annotation and evaluation guidelines are provided in \cref{appendix:guideline}.

\section{Result \& Analysis}

The performance of all models on the \textit{bangla-gloss} benchmark is summarized in \cref{table:result}. We organize our analysis around three key observations.

\paragraph{Automatic Metrics.}
GPT-5.4 achieves the highest BLEU (39.26) and chrF (73.75), establishing an upper bound among the models evaluated. Among open-source models, fine-tuned mBART attains the second-best BLEU (34.81) and chrF (70.81) despite being smaller, demonstrating the effectiveness of domain-specific fine-tuning of pretrained multilingual seq2seq models for low-resource translation. Qwen-3 follows closely with a BLEU of 30.82 and chrF of 70.16. MarianMT and Phi-4, by contrast, yield substantially lower scores, suggesting that neither a small dedicated NMT model nor a general-purpose small LM can adequately capture BdSL gloss structure under our training regime.

\paragraph{Human Evaluation.}
Human judgments reveal a notably different ranking. Qwen-3 receives the highest human evaluation score (72\%), surpassing GPT-5.4 (68\%) and mBART (65\%). This suggests that Qwen-3 produces glosses that are semantically adequate to native signers even when their surface forms diverge from the reference. MarianMT and Phi-4 receive near-zero human scores (2\% and 0\%, respectively), as both models frequently generate misspelled tokens and incoherent output.

\paragraph{Metric Discordance.}
The gap between automatic and human scores, most notably Qwen-3 ranking first in human evaluation but third in BLEU, underscores a known limitation of reference-based metrics in gloss translation: because BdSL lacks a single standardized gloss convention, valid translations may differ lexically from the reference while remaining acceptable to native signers. Our overlap analysis (\cref{app:dissimilarity}) confirms the high lexical similarity between source text and gloss (mean Jaccard Similarity: 0.5487), which further limits the discriminative power of n-gram metrics. This finding aligns with prior observations on metric reliability in low-resource sign language processing~\citep{moryossef2021data} and motivates future work on reference-free evaluation for BdSL. Illustrative model translations are in \cref{app:example}.

\subsection{Ablation Study}
To validate the contribution of each data source, we ablate on our two best-performing open-source models (Qwen-3 and mBART) under four configurations: pretrained (no fine-tuning), human-annotated only (1,000 sentences), synthetic only (1,000 subsampled), and the combined dataset (5,000 sentences). Results are in \cref{tab:ablation_results}.

\begin{table}[H]
\centering
\small
\begin{tabular}{@{}llcc@{}}
\toprule
\textbf{Model} & \textbf{Data Source} & \textbf{BLEU} & \textbf{chrF} \\
\midrule
\multirow{4}{*}{Qwen-3}
& Pretrained  & 18.85 & 58.99 \\
& Human       & 28.16 & 66.67 \\
& Synthetic   & 29.78 & 68.31 \\
& Combined    & \textbf{30.82} & \textbf{70.16} \\
\midrule
\multirow{4}{*}{mBART-large}
& Pretrained  & 23.18 & 63.72 \\
& Human       & 33.38 & 70.37 \\
& Synthetic   & 29.78 & 67.26 \\
& Combined    & \textbf{34.81} & \textbf{70.81} \\
\bottomrule
\end{tabular}
\caption{Ablation on the two best-performing open-source models: pretrained, human-only, synthetic-only (1,000 each), and combined (5,000).}
\label{tab:ablation_results}
\label{table:ablation-raw}
\end{table}

For both models, the combined dataset consistently yields the best performance. mBART improves from 23.18 (pretrained) to 33.38 (human-only) to 34.81 BLEU (combined). Qwen-3 shows a similar trend, from 18.85 to 30.82. Pretrained models perform substantially worse, confirming the necessity of domain-specific fine-tuning. Both human and synthetic data contribute complementary signal, as training on either alone yields lower scores than the combined dataset.

\section{Conclusion \& Future Work}
We presented the first parallel dataset and benchmark for Bangla text-to-gloss translation, addressing a critical gap for a sign language community of at least 3 million users. Our key finding is that a 680M-parameter mBART model, fine-tuned on just 4,000 sentences (80\% of our 5,000-sentence training set), achieves performance competitive with GPT-5.4 on automatic metrics, while Qwen-3 surpasses all models in human evaluation. This demonstrates that domain-specific adaptation is the primary driver of quality in low-resource sign language generation. Our ablation study further shows that LLM-generated synthetic data provides complementary signal to expert annotations, supporting the viability of scalable data augmentation for sign language NLP. Future work will extend the dataset with regional BdSL variations and develop reference-free evaluation metrics for this domain.

\section*{Limitations}
Gloss omits non-manual markers, so a gloss sequence accepted by a signer is a necessary but not sufficient condition for a fully faithful signed utterance. Although our dataset was annotated by people who know Bangla, it is possible that deaf people in other Bangla speaking area might follow different patterns of BdSL which are not represented in our work. Our dataset primarily consists of hand-annotated examples; however, the volume of such annotations remains limited. This constraint affects the coverage and diversity of linguistic constructions, particularly for complex or less frequently used sentence structures. Bangla exhibits unique syntactic and morphological characteristics that are reflected in Bangla Sign Language (BdSL). As such, a set of glossing rules tailored specifically to BdSL would likely produce more accurate and linguistically faithful representations. Due to the scarcity of qualified BdSL gloss annotators, we were unable to recruit a second independent evaluator for human assessment and therefore do not report inter-annotator agreement. Future work with an expanded annotator pool should address this gap.


\bibliography{custom}

\appendix

\clearpage

\section{Dataset Breakdown}
\label{app:dataset_table}

\begin{table}[htbp]
\centering
\small
\begin{tabular}{lcc}
\toprule
\textbf{Dataset} & \textbf{Train} & \textbf{Test} \\
\midrule
Human Annotated & 1,000 & 159 \\
Synthetic       & 4,000 & --  \\
\bottomrule
\end{tabular}
\caption{Dataset split across gloss experiments.}
\end{table}

\section{Existing SLT Corpora}
\label{app:sl_corpora}

\begin{table}[htbp]
\centering
\resizebox{\columnwidth}{!}{%
\begin{tabular}{@{}lcc@{}}
\toprule
\textbf{Language Pair} & \textbf{Sentences} & \textbf{Signs} \\
\midrule
Signum DGS--German~\cite{vonagris2007towards}         & 780   & 450   \\
RWTH-PHOENIX-Weather-2014T~\cite{camgoz2018neural}    & 8,257 & 1,066 \\
\bottomrule
\end{tabular}%
}
\caption{Publicly available SLT corpora with gloss annotations and spoken language translations.}
\label{tab:sl_corpora}
\end{table}

\section{Training Details}
\label{app:training}

All models were fine-tuned with Low Rank Adaptation (LoRA) on the \textit{bangla-gloss} training set. The models were trained on a single T4 GPU on Google Colaboratory. Due to resource constraints, we utilized the 4-bit quantized variant of the LLMs (specifically via Unsloth's optimized bitsAndBytes integration). This reduces memory overhead while preserving performance \cite{singhapoo2025fine}. Model-specific details are provided below.

\begin{itemize}[left=1pt]
    \item \textbf{MarianMT (Helsinki-NLP/opus-mt-en-inc):} A transformer encoder-decoder architecture trained on OPUS, a multilingual corpus of openly available documents~\cite{tiedemann2004opus}. It serves as a dedicated NMT baseline, in contrast to the general-purpose LMs listed below.
    \item \textbf{mBART-50 (facebook/mbart-large-50):} We select mBART-50 for its bidirectional Transformer architecture. Its 12-layer encoder and 12-layer decoder (with 1024-dimensional hidden states) are explicitly optimized for sequence-to-sequence reconstruction, making it superior to models that rely on separately initialized decoders such as BERT~\cite{liu2020multilingual}.
    \item \textbf{Qwen-3 (unsloth/Qwen3-8B-unsloth-bnb-4bit):} Selected for its strong multilingual performance, particularly on Indo-European languages~\cite{yang2025qwen3}. The model was fine-tuned with a maximum sequence length of 2048.
    \item \textbf{Phi-4 (unsloth/Phi-4-mini-instruct-unsloth-bnb-4bit):} A compact general-purpose language model included to test whether small LMs can match dedicated translation models or larger LLMs on the T2G task.
\end{itemize}
The values of the hyperparameter used in finetuning the models:
\begin{table}[]
    \small
\centering
    \begin{tabular}{ll}
    \toprule
    \textbf{Hyperparameter} & \textbf{Value} \\
    \midrule 
    Batch size & 8 \\ 
    Epochs & 3 \\ 
    Optimizer & AdamW (8-bit) \\ 
    Learning rate & $2 \times 10^{-5}$ \\ 
    Train/validation split & 80/20 \\ 
    Evaluation prompt & App. F \\ 
    Checkpoint Used & Final checkpoint \ (last epoch) \\
    \bottomrule
    \end{tabular}
    \caption{Details of hyperparameters used to train the models}
    \label{tab:hyperparameters}
\end{table}

\section{Sentence vs. Gloss Dissimilarity Metrics}
\label{app:dissimilarity}

To quantify the structural and lexical differences between the original spoken language sentences and their corresponding gloss versions, we employed two distinct metrics: Levenshtein Distance and Jaccard Similarity.

\textbf{Mathematical Definitions}
\begin{itemize}[left=1pt]
    \item Jaccard Similarity: Jaccard similarity is a statistical measure to quantify the similarity of two sets. If $A$ and $B$ represent the unique sets of tokens (words) in the sentence and the gloss, then Jaccard similarity, $$J(A, B) = \frac{|A \cap B|}{|A \cup B|}$$. 
    \item Levenshtein Distance: The minimum number of single-character (or word-level) edits (insertions, deletions, or substitutions) required to change one string into another. For our study, we consider character-level distance.
\end{itemize}

\begin{table}[htbp]
\centering
\small
\begin{tabular}{lcccc}
\toprule
\textbf{Metric} & \textbf{Mean} & \textbf{Median}  \\
\midrule
Jaccard Similarity & 0.5487 & --\\
Character Levenshtein Distance & 5.4088 & 4.0000  \\
Word Levenshtein Distance & 1.8931 & 2.0000 \\
Normalized Word Similarity & 0.6271 & 0.6667  \\
\bottomrule
\end{tabular}
\caption{Lexical and Structural Dissimilarity Metrics Between Sentences and Their Gloss Versions}
\label{tab:gloss_metrics}
\end{table}
From the data analysis, we observe:
\begin{enumerate}
    \item \textbf{Shared Vocabulary with Systematic Divergence:} A mean Jaccard Similarity of $0.5487$ and Normalized Word Similarity of $0.6271$ show that BdSL gloss shares substantial vocabulary with spoken Bangla, but diverges systematically. This shared vocabulary is expected, as BdSL gloss uses Bangla lexemes as labels for signs.
    \item \textbf{Morphological Transformations:} The mean Character Levenshtein Distance ($5.41$) is notably higher than the mean Word Levenshtein Distance ($1.89$), indicating that changes occur \textit{within} words rather than through word-level reordering or substitution. This pattern is consistent with morphological flattening (e.g., stripping inflections from verbs like translating \bn{খাই} into the root form \bn{খাওয়া}), which is the primary operation in BdSL glossing.
\end{enumerate}

\section{Illustrative Model Translations}
The following examples showcase model's capacity to translate Bangla text to gloss.

\begin{TranslationBox}{Example 1: Morphological Flattening (Verb Lemmatization)}
\begin{tabular}{@{}l@{\hspace{4pt}}p{0.62\linewidth}@{}}
    \textbf{Original:} & \bn{আমি বিস্কুট খাই} \\
    \textbf{English:} & I eat biscuit \\
    \textbf{Reference:}  & \bn{আমি বিস্কুট \textbf{খাওয়া}} \\
    \textbf{Predicted:} & \textcolor{successgreen}{\bn{\textbf{আমি বিস্কুট খাওয়া}}} \quad \checkmark
\end{tabular}
\end{TranslationBox}

\vspace{0.1cm}

\begin{TranslationBox}{Example 2: Particle/Punctuation Dropping}
\begin{tabular}{@{}l@{\hspace{4pt}}p{0.62\linewidth}@{}}
    \textbf{Original:} & \bn{তুমি কোথায় যাও?} \\
    \textbf{English:} & Where do you go? \\
    \textbf{Reference:}  & \bn{তুমি কোথায় যাও} \\
    \textbf{Predicted:} & \textcolor{successgreen}{\bn{\textbf{তুমি কোথায় যাও}}} \quad \checkmark
\end{tabular}
\end{TranslationBox}

\vspace{0.1cm}

\begin{TranslationBox}{Example 3: Multi-Word Syntactic Adjustment}
\begin{tabular}{@{}l@{\hspace{4pt}}p{0.62\linewidth}@{}}
    \textbf{Original:} & \bn{আমি বৃষ্টিতে ছাতা নিয়ে বাইরে যাই} \\
    \textbf{English:} & I go out in the rain with an umbrella \\
    \textbf{Reference:}  & \bn{আমি বৃষ্টি ছাতা নিয়ে আসা বাইরে যাওয়া} \\
    \textbf{Predicted:} & \textcolor{successgreen}{\textbf{\bn{আমি বৃষ্টি ছাতা নেওয়া বাইরে যাওয়া}}} \quad $\checkmark$
\end{tabular}
\end{TranslationBox}

\label{app:example}

\newtcolorbox{promptbox}[1]{
    colback=gray!5,
    colframe=gray!50,
    fonttitle=\bfseries,
    title=#1,
    sharp corners,
    boxrule=0.5pt,
    width=\linewidth,
    breakable,
    lower separated=false,
    left=5pt, right=1pt, top=5pt, bottom=5pt
}

\section{Prompt for BdSL Synthetic Gloss Generation}
\label{appendix:synth_prompt}
\begin{promptbox}{Prompt: Sentence-to-Gloss Mapping}
\raggedright\small\selectfont
You are a linguistically careful assistant trained to generate Bangla Sign Language (BdSL) gloss approximations from Bangla sentences. You are given a seed dataset with two columns: \\
\medskip
\{\{Text\}\} translates to \{\{Gloss\}\} \\
\medskip
Your task is to learn the mapping pattern between Sentence -> Gloss from the examples and then generate ONE new, high-quality gloss for ONE Bangla sentence for the topic \{\{Topics\}\}. \\
\medskip
Strict instructions: \\
1. Use Bangla words, written in Bangla script. \\
2. Use the correct infinitive form of verbs wherever applicable. Example: \bn{“কাজ করে” -> “কাজ করা”}, \bn{“গাইতে” -> “গাওয়া”}. Do not invent verb forms not seen or implied in the seed data. \\
3. Follow BdSL-preferred ordering as observed in the seed Gloss column. \\
4. Do not paraphrase or add information. \\
5. Do not remove core meaning. \\
6. If a sentence is ambiguous, choose the most neutral, literal gloss. \\
7. No numbering, no quotes, no extra whitespace. \\
8. If the input sentence contains: multiple clauses, temporal expressions, conditionals, negation, modality, or questions, the gloss MUST preserve that complexity. \\
9. Final output MUST strictly follow this format: \\
\medskip
\centering BanglaSentence -> BanglaGloss \\
\medskip
\raggedright Constraints: \\
- Exactly ONE line. \\
- Exactly ONE "->". \\
- Left side MUST be the input sentence verbatim. \\
- Right side MUST be the generated BdSL gloss. \\
- A simpler gloss than the input sentence is INVALID. \\
- Any deviation is an error.
\end{promptbox}

\section{Annotator Details}
\label{appendix:annotator}

The training dataset was annotated by a graduate student from University of Dhaka's education research institute with professional experience at a government school for the deaf. The test set was developed in collaboration with a professional sign language interpreter currently active in Bangladesh National Television. While the annotators are hearing professionals, we mitigated `hearing-centric' bias by validating our linguistic framework and findings through structured interviews with veteran BdSL educators. The primary annotator was recruited via institutional networks and was compensated at a rate exceeding local professional standards for specialized linguistic tasks, reflecting the high workload of training data preparation. The test set validator contributed their time on a voluntary basis as a professional consultant. All participants provided written informed consent. They were briefed on the project's objectives, the data's intended use for academic research, and the anonymized nature of the eventual public release. Annotators were instructed to use the natural flow of BdSL. No personally identifiable information is included in the dataset. This study was conducted under the supervision of the authors' institutional department. In the absence of a formal Institutional Review Board (IRB) for non-medical linguistic research, the protocol followed the university's internal research code of conduct to ensure the protection of all participants.

\section{Evaluation Guidelines}
\label{appendix:guideline}

\paragraph{Gloss annotation (dataset construction).}
Annotators were instructed to convert each Bangla sentence into the gloss form they would naturally use in their professional practice. The following conventions were followed:
\begin{enumerate}[left=1pt, itemsep=1pt]
    \item Verbs are reduced to their infinitive/root form (e.g., \bn{খাই} $\rightarrow$ \bn{খাওয়া}).
    \item Inflectional suffixes (case markers, tense markers) are stripped.
    \item Punctuation marks are removed.
    \item BdSL-preferred word ordering is maintained.
    \item Function words are dropped where BdSL convention omits them.
\end{enumerate}
These conventions were derived from the field study (\cref{sec:field_study}) and the government-published guide ``Isaray Bangla Bhasar Sahayika.''

\paragraph{Human evaluation (model output assessment).}
The evaluator was presented with each model-generated gloss alongside the original Bangla source sentence. The reference gloss was \textit{not} shown. For each output, the evaluator made a binary judgment: \textit{acceptable} if the generated gloss matches how the evaluator would personally sign the sentence in BdSL, or \textit{unacceptable} otherwise. The acceptability percentages reported in \cref{table:result} reflect this judgment across all 100 test sentences presented to the evaluator. We did not present them with all 159 test sentences due to time and resource constraints.

\end{document}